\documentclass[sigconf]{acmart}
\usepackage{graphicx, dblfloatfix}
\usepackage{balance}
\usepackage{amsmath,amsfonts}
\usepackage{graphicx}
\usepackage{textcomp}
\usepackage{booktabs}
\usepackage{xcolor}
\usepackage{tabularx}
\usepackage{footnote}
\usepackage{array}
\usepackage{multirow}
\usepackage{url}
\usepackage{siunitx}
\usepackage{algorithm}
\usepackage[noend]{algpseudocode}
\usepackage{dsfont}
\usepackage[bottom]{footmisc}
\usepackage{subcaption}
\usepackage{bm}
\usepackage[inline]{enumitem}

\setcopyright{acmcopyright}
\copyrightyear{2021}
\acmYear{2021}
\acmDOI{XX.XXXX/XXXXXXX.XXXXXXX}

\acmConference[SIGSPATIAL '21]{KDD '21: 27th ACM SIGKDD Conference on Knowledge Discovery and Data Mining}{November 2--5, 2021}{Beijing, China}
\acmBooktitle{29th ACM SIGSPATIAL International Conference on Advances in Geographic Information Systems,
  November 2--5, 2021, Beijing, China}
\acmPrice{15.00}
\acmISBN{XXX-X-XXXX-XXXX-X/XX/XX}

\begin{document}

\title{Auxiliary-task learning for geographic data with autoregressive embeddings}

\author{Konstantin Klemmer}
\affiliation{%
  \institution{University of Warwick \& New York University}
  \country{UK}
}
\email{k.klemmer@warwick.ac.uk}

\author{Daniel B. Neill}
\affiliation{%
  \institution{New York University}
  \country{USA}
}
\email{daniel.neill@nyu.edu}

\renewcommand{\shortauthors}{Klemmer \& Neill}

\begin{abstract}
    Machine learning is gaining popularity in a broad range of areas working with geographic data, such as ecology or atmospheric sciences. Here, data often exhibit spatial effects, which can be difficult to learn for neural networks. In this study, we propose SXL, a method for embedding information on the autoregressive nature of spatial data directly into the learning process using auxiliary tasks. We utilize the local Moran's I, a popular measure of local spatial autocorrelation, to ``nudge'' the model to learn the direction and magnitude of local spatial effects, complementing the learning of the primary task. We further introduce a novel expansion of Moran's I to multiple resolutions, thus capturing spatial interactions over longer and shorter distances simultaneously. The novel multi-resolution Moran's I can be constructed easily and as a multi-dimensional tensor offers seamless integration into existing machine learning frameworks. Throughout a range of experiments using real-world data, we highlight how our method consistently improves the training of neural networks in unsupervised and supervised learning tasks. In generative spatial modeling experiments, we propose a novel loss for auxiliary task GANs utilizing task uncertainty weights. Our proposed method outperforms domain-specific spatial interpolation benchmarks, highlighting its potential for downstream applications. This study bridges expertise from geographic information science and machine learning, showing how this integration of disciplines can help to address domain-specific challenges. The code for our experiments is available via \url{https://drive.google.com/file/d/1ShOBV7RifMdS9LYsySOM084KL5aD-wkm/view?usp=sharing}.
\end{abstract}



\maketitle

\section{Introduction}

When monitoring the physical environment, from crop yields to air pollution, the gathered data are often geographic in nature and follow some spatial process: data values depend on their spatial locations. This violates a key assumption of many statistical learning frameworks, that data are independent and identically distributed (\emph{iid}). Even models which are explicitly designed to account for these dependencies, such as kernel methods, may struggle with over- and under-smoothing of spatial patterns or with distinguishing stationary from non-stationary spatial effects. The complexities of geographic data concern researchers in many academic fields and are a key focus of the geographic information sciences (GIS). The GIS community has a long tradition of analyzing spatial phenomena, developing metrics to measure spatial effects and deploying models to account for spatial dependencies \cite{Cressie1991, Anselin1988, Lichstein2002}. With the growing popularity of scalable machine learning methods, particularly deep neural networks, applications of these approaches to geospatial data domains have become more and more common. Nevertheless, these applications have only rarely inspired methodological innovation \cite{Janowicz2020} in neural networks. Recently, \cite{Reichstein2019} specifically pointed to a lack of deep learning methods tailored to geospatial and spatio-temporal data in the context of earth system science. In light of this call-to-action, we propose \texttt{SXL}, a novel neural network method for geospatial data domains, that is inspired by domain expertise from GIS and explicitly learns spatial dependencies contained within the data. This is facilitated through a multi-task learning process, where a spatial embedding capturing local autoregressive effects at each data point is learned as an auxiliary task, sharing the model parameters with the main task and hence nudging the model to learn both the original data and its embedding in parallel. We integrate one of the most prominent and widely used metrics in GIS into the model: the Moran's I measure of local spatial autocorrelation \cite{Anselin1995}. To also account for longer-distance spatial relations, we propose a novel multi-resolution local Moran's I by gradually coarsening the input data. By providing a learner with prior knowledge on the autoregressive nature of the data we seek to improve performance on a broad range of spatial modeling tasks. 

Our main contributions can be summarized as follows: \textit{(1)} We propose a novel, flexible multi-resolution expansion of the Moran's I measure of local spatial autocorrelation. \textit{2)} We use the traditional and multi-resolution local Moran's I as embeddings to be learned in an auxiliary learning procedure to capture both short- and long-distance spatial effects. \textit{(3)} We provide a practical framework to adapt our method to arbitrary neural network architectures and different supervised (predictive) and unsupervised (generative) spatial learning tasks. \textit{(4)} For the purpose of balancing the losses of multiple tasks in a generative modeling setting, we develop a novel auxiliary task GAN loss based on uncertainty weights \cite{Cipolla2018}. \textit{(5)} We evaluate \texttt{SXL} on both generative and predictive spatial modeling tasks, providing empirical evidence for consistent and robust performance gains across multiple synthetic and real-world experimental settings. 
\section{Related work}

\subsection{Machine learning with geospatial data}

Geography has a long tradition of modeling and empirical analysis. Ideas from GIS and spatial statistics have inspired popular approaches in modern machine learning, from Gaussian Processes (GPs), pioneered by the development of Kriging in the 1960s \citep{Matheron1963}, to spatial scan statistics\citep{Kulldorff1997}, widely used for event and pattern detection. Nevertheless, since the emergence of the era of deep neural networks, the relationship between the GIS and machine learning communities have been mostly defined through applications of existing neural network models to geographic data. Only rarely have concepts and ideas from GIS and spatial statistics motivated methodological advancements in neural networks--whereas application areas such as vision, bioinformatics, and computational linguistics have strongly influenced the deep learning state-of-the-art we know today. Nevertheless, some recent advances in machine learning have been exceptionally useful to the GIS community. Kernel methods such as GPs have seen huge progress towards overcoming computational bottlenecks stemming from the complexities of working with pair-wise distance matrices. Kernel interpolation techniques and GPU acceleration have allowed GPs to easily scale up to a million data points \citep{Gardner2018,Wang2019a}. In neural networks, the emergence of graph-based networks and particularly graph convolutions have allowed for the modeling of asymmetric and non-Euclidean spatial relationships \citep{Li2019}, and the emergence of physics-informed deep learning has reinforced the need for neural networks to model complex spatio-temporal patterns \citep{Wang2020}.

Nevertheless, as geographic context has proven to be of relevance in many machine learning applications \citep{Yan2017,Aodha2019,Chu2019,Bao2020}, core concepts from GIS and geography have gradually attracted more attention in the ML community. While methodologies for geospatial data and problems have previously utilized traditional machine learning approaches like tree-based models \cite{Jiang2012} or ensemble learners \citep{Jiang2017}, focus has recently switched more towards neural networks: for example, \citep{Mai2020} and \citep{Yin2019} propose a vector embedding to capture spatial context, inspired by word embeddings such as \textit{Word2Vec} \citep{Mikolov2013}. \citep{Zammit-Mangion2019} propose to model complex spatial covariance patterns through injective warping functions, learned by a deep net. While there exist approaches for capturing spatial autocorrelation in neural networks \citep{Zhang2019}, the Moran's I metric has previously only been used once in an explicit machine learning setting: \citep{Klemmer2019a} propose to use it as a heuristic for early stopping in generative adversarial network (GAN) training, but not explicitly for learning. More generally, geographic metrics describing spatial phenomena have barely been used in neural network models--which we believe is a missed opportunity.

\subsection{Learning with auxiliary tasks}

In this paper, we use auxiliary learning, an approach using multi-task learning to improve performance on a primary task, originally conceptualized by \citep{Suddarth1990}. The authors propose to give learners ``hints'' related to the original task throughout training in order to improve training speed and model performance. This can be understood as forcing the learner (e.g., a neural network) to focus its attention on certain patterns in the data, highlighted by the auxiliary objective. It also implies that the auxiliary task has to provide some meaningful embedding of the primary task. Auxiliary learning has been particularly successful in deep reinforcement learning and is widely utilized \citep{Flet-Berliac2019}: for example, learning to steer a wheel can be improved using auxiliary tasks related to image segmentation and optical flow estimation \citep{Hou2019}. These auxiliary tasks are already related to spatial perception, which is common when working with image data: recent work has highlighted the applicability of pixel control tasks \citep{Jaderberg2017} or depth estimation  \citep{Mordan2018}. Auxiliary tasks have also been successfully deployed for generative modeling with GANs: Auxiliary Classifier GAN (AC-GAN) \cite{Odena2017} extends the original GAN loss function by an auxiliary classifier to improve the fidelity of generated images. However, this approach comes with complications. As recent work \citep{Gong2019} notes, AC-GANs can lead to perfect separability--where the GAN discriminator is easily winning the two-player game against the generator, preventing efficient and balanced learning. Another use of spatial semantics in an auxiliary task GAN setting is proposed by \citep{Wang2016c}: the authors propose to generate surface normal maps, a 3D representation of the 2D image, and use these in an auxiliary task. In an explicit geospatial setting, auxiliary tasks have been used in semantic visual localization \citep{Schonberger2018} and semantic segmentation \citep{Chatterjee2019}, in addition to dedicated spatial and spatio-temporal multi-task frameworks \citep{Xu2016}. Nevertheless, to the best of our knowledge, measures of spatial autocorrelation such as Moran's I have never previously been used in any kind of multi-task learning setting, be it for generative or predictive spatial modeling.

\section{Methodology}

\subsection{Multi-resolution local Moran's I}

\begin{figure}[!htbp]
\centering
\includegraphics[scale=0.5]{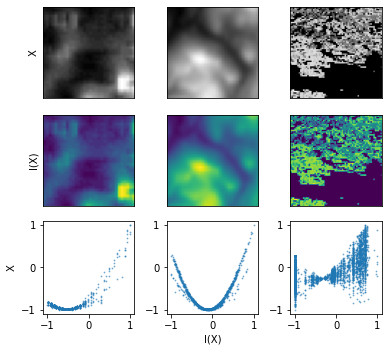}
\caption{Example of the different shapes the relationship between $X$ and $I(X)$ can take using seabed relief, digital elevation map (DEM) and tree canopy data (from left to right). The figure also highlights how the local Moran's I can work both, as a measure of spatial outliers (first column) and homogenous spatial clusters (third column).}
\label{figm}
\end{figure}

\begin{figure*}[htpb]
\centering
\includegraphics[scale=0.3]{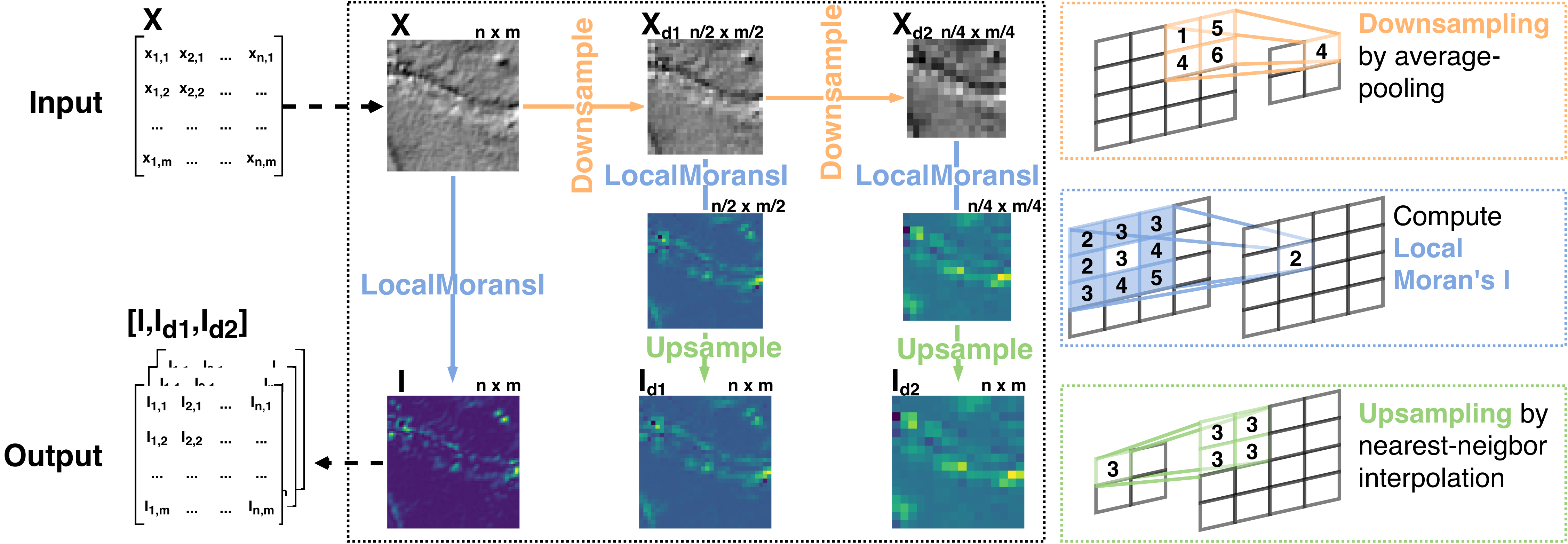}
\caption{Multi-resolution local Moran's I calculation with an example input at three different resolutions: Original input size ($n \times m$), downsampled by factor $2$ ($n/2 \times m/2$) and downsampled by factor $4$ ($n/4 \times m/4$).}
\label{fig1}
\vskip -0.15in
\end{figure*}

As outlined above, working with geospatial data requires a careful assessment of and accounting for potential autoregressive effects--an intuition which neural networks traditionally do not provide. One of the most prominent measures, capturing spatial autocorrelation at the point-level, is the local Moran's I metric \citep{Anselin1995}. Local Moran's I measures the direction and extent of similarity between each observation and its local spatial neighbourhood. As such, it provides an indication for both local spatial clusters and spatial outliers.

Formally, let $X$ be a 2-$d$ spatial matrix (image)
\begin{equation}
    X^{n \times m} = 
        \begin{bmatrix}
            x_{1,1} & \dots & x_{1,m} \\
            \dots & \dots & \dots \\
            x_{n,1} & \dots & x_{n,m}
        \end{bmatrix}
\end{equation}

and the vector $\mathbf{x} = vec(X)$ consists of $n_{x} = nm$  real-valued observations $x_{i}$, referenced by an index set $N_{x} = \{1,2,...,n_{x} \}$. We define the \textit{spatial neighbourhood} of observation $i$ to be $\mathcal{N}_{x_{i}} = \{ j \in N_{x} : w_{i,j} > 0 \}$. Here, $w_{i,j}$ corresponds to a binary spatial weight matrix, indicating whether any observation $j$ is a neighbor of $i$. For continuous data, the creation of this matrix requires computing a pair-wise distance matrix or kd-tree, however in our case using discrete $n \times m$ matrices, this problem is trivial. Throughout this study, we utilize queen contiguity (i.e. all adjacent grid cells, including diagonals, are neighbors), but the approach generalizes to arbitrary neighborhood definitions. We can compute the local Moran's I statistic, $I_{i}$, of observation $x_{i}$ with the mean over all observations $\bar{x}$ as:

\begin{equation}
\begin{split}
    I_{i} = (n_{x}-1) \frac{x_{i} - \bar{x}}{\sum^{n_{x}}_{j=1} (x_{j} - \bar{x})^2} \sum^{n_x}_{j=1, j \neq i} w_{i,j} (x_{j} - \bar{x})
\end{split}
\end{equation}
Combining all local Moran's I values gives the matrix $I^{n \times m}$, of the same size as $X^{n \times m}$. $I_{i}$ can take positive or negative values: a positive value suggests that a data point is similar to its neighbors, which could indicate latent cluster structure. A negative value suggests that the data point is distinctly different from neighboring data points, which could indicate a changepoint or edge. While the Local Moran's I statistic is closely correlated to its input, their relationship can take different forms, depending on the complexity of the inputs' spatial structure. Figure 1 provides examples of different geospatial data $X$ and their respective $I(X)$ representations. 

One of the main limitations of the local Moran's I metric is its restriction to represent local spatial dependence only at the scale provided by the immediate neighbors as defined by the neighborhood matrix $w$. Thus longer-range spatial dependencies can be lost. This scale sensitivity of the local Moran's I is known as a common challenge in applications \citep{Feng2019,Meng2014,Zhang2011}. Nevertheless, to our knowledge, no alternative metric accounting for this issue exists. Here, we propose a novel, multi-resolution representation of the local Moran's I by increasingly coarsening the input data for the Moran's I computation and then upsampling the output back to the original data shape. The coarsening step here is analogous to a 2-$d$ average pooling operation, so that our coarsened input is given as:

\begin{equation}
\begin{split}
    X^{n/a \times m/a}_{d}(i,j)  = \text{mean}\{ X^{n \times m}(ai + k, aj + l)\}, \: \\
    \text{for} \: 0 \leq k < a \; \text{and} \; 0 \leq l < a,
\end{split}
\end{equation}

where $a$ gives the kernel size and the subscript $d$ indicates \textit{downsampling}. The coarsened spatial matrix $X_{d}$ corresponds to the vector $\mathbf{x}^{(d)}$ of length $n_{x^{(d)}}$. The coarsened local Moran's I, $I(\mathbf{x}^{(d)})$, is then computed according to Equation (2), using the spatial weight matrix $w^{x^{(d)}}$, corresponding to the new size $n/a \times m/a$ of the  downsampled input. In the last step, the coarsened Moran's I is upsampled again to the original input size $n \times m$ using nearest-neighbor interpolation. This whole process can be repeated several times to compute the local Moran's I at increasingly coarse resolutions. The local Moran's I values at different resolutions can then be stacked on top of one another, much like a multi-channel image (e.g. RGB image). As such, tensors provide an ideal data structure for our metric. We illustrate this with an example in Figure 2.

\subsection{Auxiliary learning of spatial autoregressive structures}

Auxiliary task learning shares the benefits of multi-task learning: auxiliary tasks hint at specific patterns in the data for the model to focus attention on. Further, they introduce a representation bias, whereas the model prefers latent representations of the data that work for both primary and auxiliary tasks, thus helping with generalization. Lastly, auxiliary tasks can work as regularizers by introducing inductive bias and decreasing the risk of overfitting the model. Here, we want to use the local Moran's I embedding and our newly introduced multi-resolution Moran's I as auxiliary tasks. The main motivation for any auxiliary tasks is ``relatedness'' to the primary task: spatial theory characterizes a spatial pattern as a reflection of underlying spatial processes. Accordingly, \citep{Chou1995} concludes that ``[...] the capability of generalizing and quantifying spatial patterns is a prerequisite to understanding the complicated processes governing the distribution of spatial phenomena.''--explicitly mentioning the power of the Moran's I metric to capture these effects. This statement can be translated directly into a learning algorithm, where the learning of a spatial pattern is constrained by a simultaneous learning of the underlying spatial process. Together with the well documented success of spatial auxiliary tasks in computer vision (see \textit{Related Work}), this makes auxiliary tasks based on the local Moran's I well motivated by both spatial theory and machine learning research. Recent research further highlights the importance of learning at multiple resolutions to support a comprehensive understanding of spatial processes \citep{Park2020,Tanaka2019,Xu2018}. Lastly, the local Moran's I (and the multi-resolution variant) can be constructed for any numerical input, and can thus be seamlessly integrated in a broad range of learning settings and with arbitrary neural network architectures. 

With our experiments, we focus on two distinct settings: generative spatial modeling using GANs \citep{Goodfellow2014}, and predictive spatial modeling in the form of spatial interpolation. To outline the application of our proposed auxiliary task approach, we introduce the GAN example in detail--the predictive modeling formulation follows from this straightforward. GANs are a family of generative models comprised of two neural networks, a Generator \textbf{G} that produces fake data and a Discriminator \textbf{D} that seeks to distinguish between real and fake data. These two networks are agents in a two-player game, where \textbf{G} learns to produce synthetic data samples that are faithful to the true data generating process, and \textbf{D} learns to separate real from fake samples, thus pushing \textbf{G} to produce increasingly realistic synthetic data. The standard GAN loss function is thus given as:

\begin{equation}
\begin{aligned}
    \min_G \max_D \mathcal{L}_{GAN} (D,G) = \mathbb{E}_{\mathbf{x} \sim p_{data}(\mathbf{x})} \bigl[\log D(\mathbf{x})\bigr] + \\
    \mathbb{E}_{\mathbf{z} \sim p_{\mathbf{z}}(\mathbf{z})} \bigl[\log (1 - D(G(\mathbf{z})))\bigr],
\end{aligned}    
\end{equation}

\noindent consisting of the Discriminator and Generator losses

\begin{equation}
    \mathcal{L}^{(D)}_{GAN} = \max_D [\log(D(\mathbf{x})) + \log(1-D(G(\mathbf{z})))],
\end{equation}
\begin{equation}
    \mathcal{L}^{(G)}_{GAN} = \min_G [\log(D(\mathbf{x})) + \log(1-D(G(\mathbf{z})))].
\end{equation}

Our auxiliary task approach augments the Discriminator with a loss based on the Moran's I embeddings of the real and the fake data:

\begin{equation}
    \mathcal{L}^{(D)}_{AT} = \max_D [\log(D(I(\mathbf{x}))) + \log(1-D(I(G(\mathbf{z}))))],
\end{equation}

\noindent so that the composite loss for $N$ auxiliary tasks (single- or multi-resolution) is given as:

\begin{equation}
\begin{aligned}
    \min_G \max_D \mathcal{L}_{MRES-MAT} (D,G) = \mathcal{L}_{GAN} (D,G) + \\
    \lambda (\mathcal{L}_{AT_{1}}^{(D)} + \dots + \mathcal{L}_{AT_{N}}^{(D)}).
\end{aligned}
\end{equation}

Both loss functions use a customary weight hyper-parameter $\lambda$ for the auxiliary losses. Alternatively, we could fit separate weights for each auxiliary task. The approaches are further illustrated in Figure 3. Integrating the auxiliary tasks into predictive models for spatial interpolation is more straightforward: We simply let a regressor $f$ predict the (multi-resolution) local Moran's I of the output, $I(y) \sim f(I(x))$, simultaneously with the main task $y \sim f(x)$. 

\subsection{Loss balancing using task uncertainty}

\begin{figure*}
\centering
\includegraphics[scale=0.44]{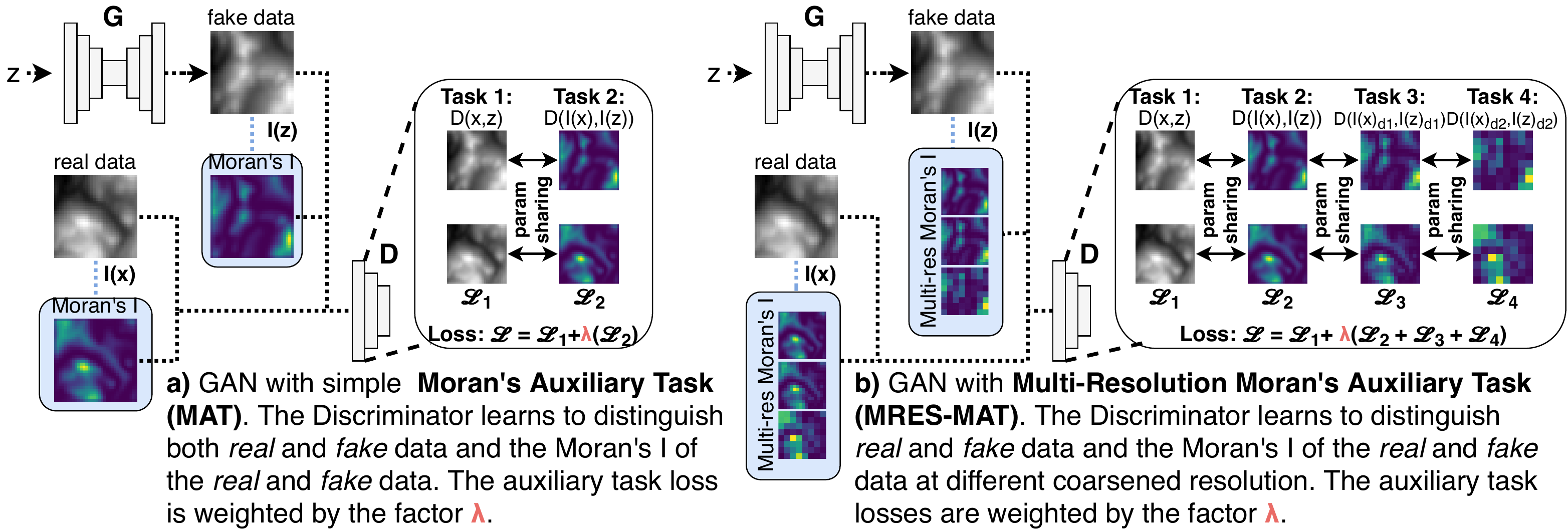}
\caption{GAN with spatially explicit auxiliary tasks, using the Moran's I (\textbf{a}) and multi-resolution Moran's I (\textbf{b}) embeddings in the Discriminator.}
\label{fig2}
\end{figure*}

In addition to the hard loss weight $\lambda$ outlined above, we also test a setting for automatically learning the loss weight of main and auxiliary tasks. For this, we follow the approach conceptualized by \citep{Cipolla2018} and utilize each task's homoskedastic uncertainty to inform the loss weight. To formalize this, \citep{Cipolla2018} define a probabilistic multi-task regression problem with $N$ tasks and likelihood

\begin{equation}
\begin{aligned}
    p(\mathbf{y}_1,\dots,\mathbf{y}_N | f(\mathbf{x})) = p(\mathbf{y}_1 | f(\mathbf{x})) \dots p(\mathbf{y}_N | f(\mathbf{x})),
\end{aligned}
\end{equation}

\noindent where $\mathbf{y}_1,\dots,\mathbf{y}_N$ are the main and auxiliary model outputs and $\mathbf{x}$ is the model input. Using the maximum likelihood method, the minimization objective of the multi-task regression is $\min \mathcal{L}(\sigma_1,\dots,\sigma_N)$:

\begin{equation}
\begin{aligned}
    & = -\log p(\mathbf{y}_1,...,\mathbf{y}_N | f(\mathbf{x})) \\
    & = \frac{1}{2 \sigma_{1}^{2}} \mathcal{L}_{1} + \dots + \frac{1}{2 \sigma_{N}^{2}} \mathcal{L}_{N} + \sum_{i=1}^{N} \log \sigma_i, 
\end{aligned}
\end{equation}

\noindent where $\sigma_1,\dots,\sigma_N$ are the model noise parameters. Minimizing this objective can be interpreted as learning the relative weight of each task's contribution to the composite loss. The noise is kept from decreasing infinitely by the last term of the loss, which serves as a regularizer. This loss constitutes the objective we use for our predictive modeling task. However, no approach for uncertainty weighted auxiliary task GANs exists. We hence propose an extension to the auxiliary GAN loss outlined in the previous section: Instead of using a fixed auxiliary task weight $\lambda$, we augment both main and auxiliary discriminator losses with uncertainty weights, so that

\begin{equation}
\begin{aligned}
    \min_G \max_D \mathcal{L}_{MRES-MAT} (D,G) = \mathcal{L}_{GAN}^{(G)} + \\
    (\frac{1}{2 \sigma_{1}^{2}} \mathcal{L}_{AT_{1}}^{(D)} + \dots + \frac{1}{2 \sigma_{N}^{2}} \mathcal{L}_{AT_{N}}^{(D)} + \sum_{i=1}^{N} \log \sigma_i).
\end{aligned}
\end{equation}

This constitutes the first adaption of the uncertainty task balancing principles to the multi-task GAN family. In the following experiments, we report results from both fixed weight parameters $\lambda$ and uncertainty weights. Throughout rigorous experiments, we find weights of $\lambda = [0.1, 0.01]$ to be particularly helpful and hence utilize these throughout our experiments. We use auxiliary learning settings with hard-parameter sharing, where the top layers of the respective models are task-specific. For the GAN experiment, just the last layer is task-specific, and for the interpolation experiment the last two layers are task specific.

Having outlined our approach for GANs in detail, the adaptation of \textit{SXL} to predictive spatial models is trivial: Just as with the GAN discriminator, a second prediction head is added to the predictive model (e.g. a neural network regressor for spatial interpolation), aiming to predict the (multi-resolution) local Moran's I embedding of the output parallely to the main task. The losses of both tasks are then combined the same way as for the GAN discriminator, using either a hard weighting parameter or uncertainty weights, as outlined in \textit{Equation 10}. 
\section{Experiments}

\subsection{\textit{Experiment 1:} Generative spatial modeling}

\begin{figure*}[!ht]
    \centering
    \begin{subfigure}{.33\textwidth}
      \centering
      \includegraphics[scale=0.4]{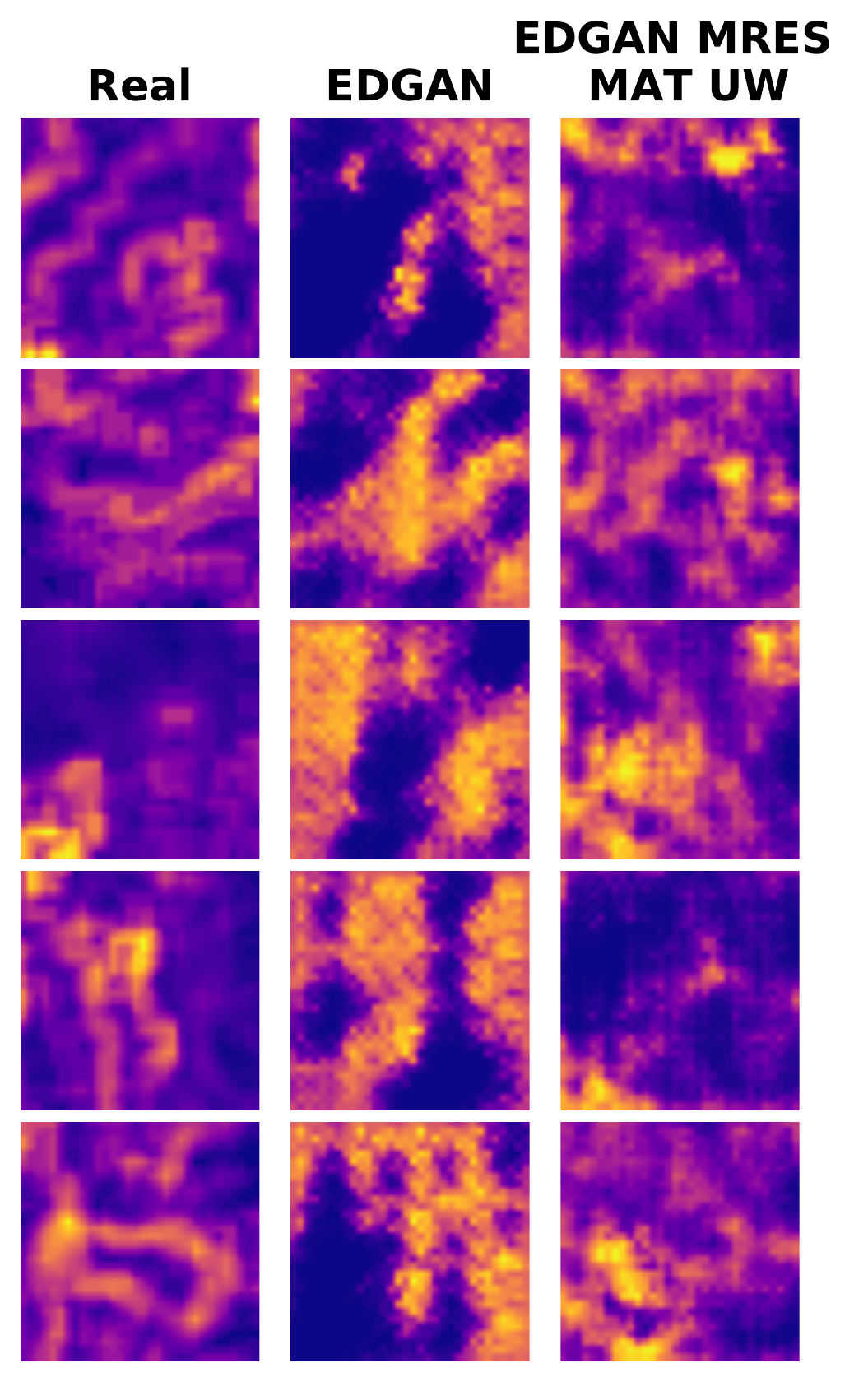}
      \caption{\textit{PetrelGrid}: Smoothing out \\\hspace{\textwidth} edges and spatial changepoints.}
      \label{fig:sub1}
    \end{subfigure}%
    \begin{subfigure}{.33\textwidth}
      \centering
      \includegraphics[scale=0.4]{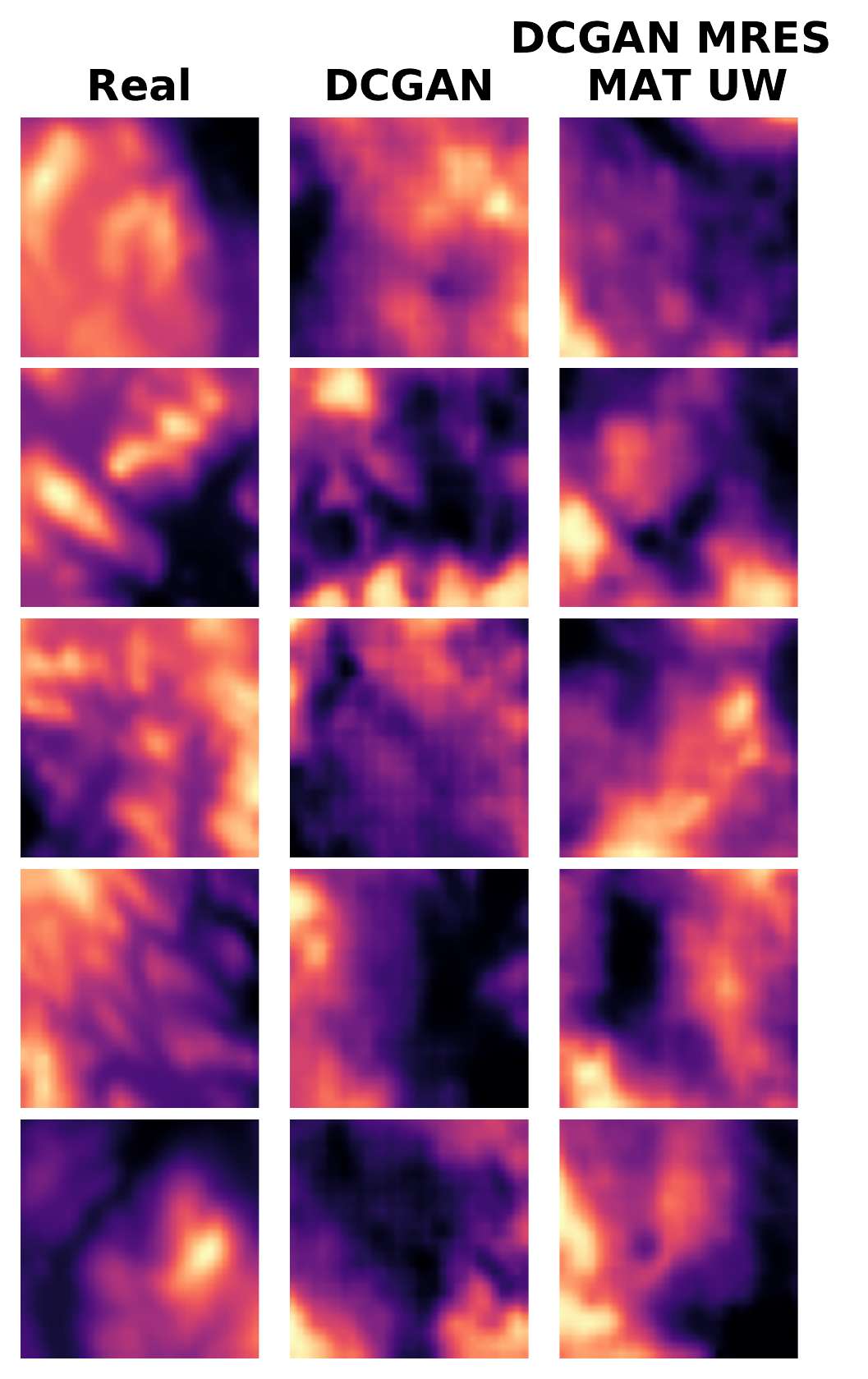}
      \caption{\textit{DEM}: Reducing \\\hspace{\textwidth} checkerboard artifacts.}
      \label{fig:sub1}
    \end{subfigure}%
    \begin{subfigure}{.33\textwidth}
      \centering
      \includegraphics[scale=0.4]{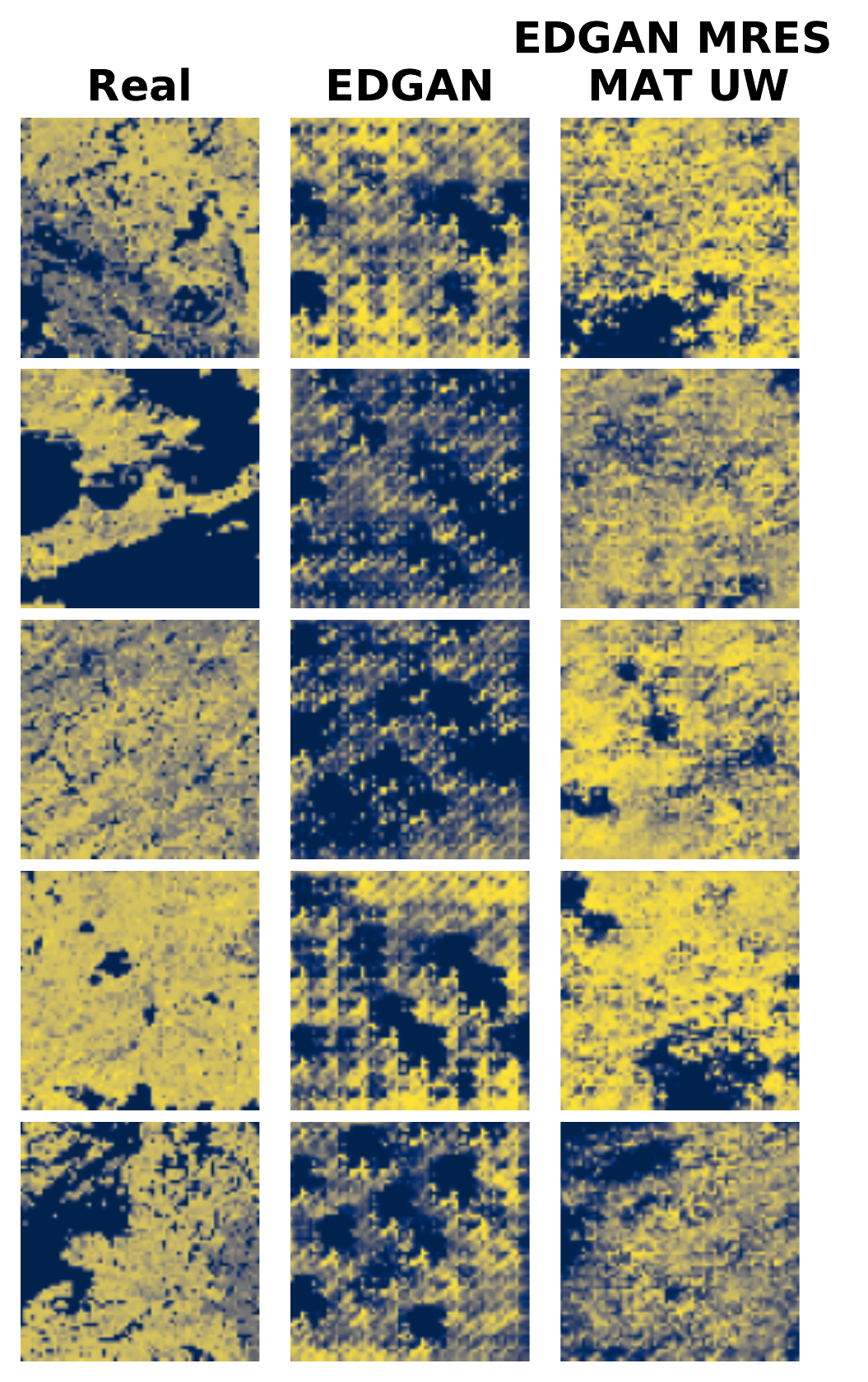}
      \caption{\textit{TreeCanopy}: Reducing \\\hspace{\textwidth} (partial) mode collapse.}
      \label{fig:sub2}
    \end{subfigure}
    \caption{Example images highlighting the positive effects of \textit{MAT} / \textit{MRES MAT} on different GAN architectures, across different datasets.}
    \label{fig:test}
\end{figure*}

\begin{table*}[!ht]
\centering
\resizebox{11cm}{!}{%
\begin{tabular}{l|l|l|l|l}
\textbf{Model} & \textbf{Toy} & \textbf{PetrelGrid} & \textbf{DEM} & \textbf{TreeCanopy} \\		
\hline
\hline
GAN \citep{Goodfellow2014}	& $	0.0934	$ & $	0.4106	$ & $	\mathbf{0.1120}	$ & $	0.1138	$ \\
GAN-MAT UW	& $	0.1077	$ & $	0.4860	$ & $	0.1814	$ & $	0.1132	$ \\
GAN-MRES-MAT UW	& $	\mathbf{0.0917}	$ & $	\mathbf{0.4014}	$ & $	0.1180	$ & $	\mathbf{0.1038}	$ \\
\hline
DCGAN \citep{Radford2016}	& $	0.1534	$ & $	0.2993	$ & $	0.0591	$ & $	0.0654	$ \\
DCGAN-MAT UW	& $	0.2319	$ & $	0.3049	$ & $	\mathbf{0.0591}	$ & $	0.1009	$ \\
DCGAN-MRES-MAT UW	& $	\mathbf{0.0938}	$ & $	\mathbf{0.2793}	$ & $	0.0612	$ & $	\mathbf{0.0635}	$ \\
\hline
EDGAN \citep{Zhu2019}	& $	0.0269	$ & $	\mathbf{0.2909}	$ & $	0.0499	$ & $	0.0322	$ \\
EDGAN-MAT UW	& $	0.0276	$ & $	0.3061	$ & $	0.0481	$ & $	0.0316	$ \\
EDGAN-MRES-MAT UW	& $	\mathbf{0.0241}	$ & $	0.2971	$ & $	\mathbf{0.0469}	$ & $	\mathbf{0.0314}	$
\end{tabular}}
\caption{Test MMD scores (lower is better) of different GAN configurations. We compare synthetic samples from these generators to held-out test data to compute the scores. Shown are models trained with uncertainty weighted auxiliary tasks.} \label{tab:table2}
\end{table*}

\begin{table*}[!ht]
\centering
\resizebox{11cm}{!}{%
\begin{tabular}{l|l|l|l|l}
\textbf{Model} & \textbf{Toy} & \textbf{PetrelGrid} & \textbf{DEM} & \textbf{TreeCanopy} \\		
\hline
\hline
GAN \citep{Goodfellow2014}	& $	0.0934	$ & $	0.4106	$ & $	0.1120	$ & $	0.1138	$ \\
GAN-MAT $\lambda = 0.1$	& $	0.0994	$ & $	0.4319	$ & $	0.1028	$ & $	0.1140	$ \\
GAN-MAT $\lambda = 0.01$	& $	0.1125	$ & $	\mathbf{0.3965}	$ & $	0.1034	$ & $	\mathbf{0.0972}	$ \\
GAN-MRES-MAT $\lambda = 0.1$	& $	\mathbf{0.0922}	$ & $	0.4394	$ & $	0.1086	$ & $	0.1221	$  \\
GAN-MRES-MAT $\lambda = 0.01$	& $	0.1133	$ & $	0.3989	$ & $	\mathbf{0.0983}	$ & $	0.1026	$ \\
\hline
DCGAN \citep{Arjovsky2017}	& $	0.1534	$ & $	0.2993	$ & $	0.0591	$ & $	0.0654	$ \\
DCGAN-MAT $\lambda = 0.1$	& $	0.2360	$ & $	0.2858	$ & $	\mathbf{0.0569}	$ & $	0.0692	$ \\
DCGAN-MAT $\lambda = 0.01$	& $	\mathbf{0.1494}	$ & $	0.2977	$ & $	0.0578	$ & $	\mathbf{0.0596}	$ \\
DCGAN-MRES-MAT $\lambda = 0.1$	& $	0.2147	$ & $	\mathbf{0.2828}	$ & $	0.0590	$ & $	0.0602	$ \\
DCGAN-MRES-MAT $\lambda = 0.01$	& $	0.2154	$ & $	0.2868	$ & $	0.0576	$ & $	0.0640	$ \\
\hline
EDGAN \citep{Zhu2019}	& $	0.0269	$ & $	0.2909	$ & $	0.0499	$ & $	0.0322	$ \\
EDGAN-MAT $\lambda = 0.1$	& $	0.0289	$ & $	0.2973	$ & $	0.0460	$ & $	\mathbf{0.0316}	$ \\
EDGAN-MAT $\lambda = 0.01$	& $	0.0269	$ & $	0.3012	$ & $	0.0467	$ & $	0.0319	$	\\
EDGAN-MRES-MAT $\lambda = 0.1$	& $	0.0253	$ & $	\mathbf{0.2676}	$ & $	0.0482	$ & $	0.0317	$ \\
EDGAN-MRES-MAT $\lambda = 0.01$	& $	\mathbf{0.0247}	$ & $	0.2898	$ & $	\mathbf{0.0438}	$ & $	0.0321$

\end{tabular}}
\caption{Test MMD scores (lower is better) of different GAN architectures. We compare synthetic samples from these generators to held-out test data to compute the scores. Shown are models trained with hard auxiliary task weights $\lambda$.} \label{tab:table9}
\end{table*}

\textbf{\textit{Selection \& Evaluation}}: In our first experiment, we want to examine whether our proposed method can improve the learning of spatial data generating processes. We generate synthetic data from several types of GANs, with and without including the Moran's I auxiliary tasks, and compare how faithful the generated samples are compared to the true distribution of samples. To assess model quality, we use the Maximum Mean Discrepancy (MMD) metric \citep{Borgwardt2006}, a distance measure between distributions based on mean embeddings of the features. For data distributions $P$ and $Q$, the MMD is defined as $MMD(P,Q) = || \mu_{P} - \mu_{Q} ||_{\mathbb{R}^{d}}$. The empirical MMD for random variables $x_{i}$ and $y_{i}$ of length $n$ is given as

\begin{equation}
\begin{aligned}
    \widehat{MMD}^{2} = \frac{1} {n(n - 1)} \sum_{i \neq j} k(x_{i},x_{j}) +  \\
    \frac{1}{n(n - 1)} \sum_{i \neq j} k(y_{i},y_{j}) -  \frac{2}{n^{2}} \sum_{i,j} k(x_{i},y_{j}),
\end{aligned}
\end{equation}

where $k: \mathcal{X} \times \mathcal{X}$ represents a positive-definite kernel---in our case a radial basis function (RBF) kernel. The more similar the data distributions $P$ and $Q$ are, the closer the MMD metric gets to $0$. A lower MMD score between samples of real and synthetic data thus indicates higher quality of the synthetic samples. As GAN training is notoriously difficult and prone to mode collapse and other issues, we opt for the following training and selection procedure to ensure our findings are robust: For each architecture and training strategy combination (e.g. EDGAN \textbf{MRES MAT UW}) we train ten cycles of GANs. For each cycle, we save the  generator which optimizes the MMD metric on a held-out validation set (separate from the training data and from the test data used for evaluation), rather than choosing the final model after all training iterations. We then choose the one of these ten generators that optimizes the MMD metric on the validation set, and finally evaluate the MMD score of that generator on a separate, held-out test set. In short, this training process allows each architecture and training strategy combination ten cycles to train the best possible generator (as measured on the validation set), which is then evaluated on the test set and compared to all other combinations of architecture and training strategy. 

\textbf{\textit{Data}}: We select four datasets for our experiments: \textit{(1)} A toy dataset of $7000$ $32 \times 32$ tiles with a Gaussian peak and a Gaussian dip, where the position of the dip mirrors the position of the peak. \textit{(2)} The \textit{PetrelGrid} seabed relief dataset \citep{Li2013}, processed into a grid of $195$ $32 \times 32$ tiles. \textit{(3)} Digital Elevation Model (\textit{DEM}) data of the area surrounding Lake Sunapee (NH, USA), as found in the \texttt{elevatr} \textit{R} package \footnote{See: \url{https://github.com/jhollist/elevatr}}, processed into a grid of $1156$ $32 \times 32$ tiles. \textit{(4)} \textit{Tree canopy} data of the University of Maryland's ``Global Tree Change'' project \citep{Hansen2013}, processed into a grid of $1800$ $64 \times 64$ tiles. These datasets are chosen to represent a range of different geospatial patterns occurring in real-world physical environments and relate to important modeling challenges in the earth sciences, ecology, or geography. For more information on the data used, including summary statistics, please refer to Appendix \textit{A}.

\textbf{\textit{Benchmark Models}}: The modularity of our proposed auxiliary task learning method allows us to test it on a range of different GAN models. We chose the original GAN implementation, denoted here as \textit{GAN}~\citep{Goodfellow2014}, the \textit{DCGAN}~\cite{Radford2016} and lastly an Encoder-Decoder GAN (\textit{EDGAN}) architecture recently proposed by~\citep{Zhu2019} and explicitly designed for geospatial applications. All models are optimized using the same, traditional GAN objective lined out in the previous section. We test all benchmark models with the single- and multi-resolution Moran's I auxiliary task (\textbf{MAT} / \textbf{MRES-MAT}) as well as with fixed ($\lambda$) and uncertainty (\textbf{UW}) based task weights. For more details on the experimental setup, including hyperparameters and detailed model architectures, please refer to Appendix \textit{B.I}.


\textbf{\textit{Findings}}: Table 1 and 2 show the MMD scores of generators selected according to the strategy outlined in the \textit{Evaluation \& Selection} paragraph. Table 1 highlights results from the uncertainty weighting strategy, and Table 2 from the hard loss weights $\lambda$. We can see that for both strategies the auxiliary task settings improve performance for most experiments, agnostic of the underlying GAN architecture, by usually 3-10\%. We believe the auxiliary tasks support the learning process in two ways: (1) GAN Discriminators are known to exhibit spare capacity (i.e., they can be too powerful), which can cause them to over-specialize, leading to worse generalization performance \citep{Hardy2019}---thus adding a second, closely correlated task should not pose a problem but might help prevent over-specialization in the bottom Discriminator layers. (2) The losses stemming from the auxiliary tasks have a regularizing effect throughout training, further preventing Discriminator over-specialization. 

This leads to several beneficial effects, some even visually apparent, as highlighted in Figure 4. Figure 4a shows how generators trained with \textbf{MRES MAT} appear to be better at smoothing (but not over=smoothing) spatial artifacts---residual, noisy spatial patterns and hard edges introduced by the model---compared to the same GAN backbone trained without our auxiliary task. Figure 4c shows two EDGANs, one trained with and one trained without \textbf{MRES-MAT}. We observe that the model trained without the auxiliary task exhibits ``mode collapse'', a phenomenon common with GANs where a Generator always produces the same image or some parts of the image are always the same, while the Generator including the auxiliary task does not exhibit this behavior. It is important to note that these examples are not cherry-picked but represent a pattern we can observe throughout all our experiments: the auxiliary tasks consistently improved model performance. 

However, the optimal setting for applying the auxiliary task appears to vary, both in terms of the task weighting strategy and using the simple \textbf{MAT} versus the \textbf{MRES-MAT}. For example, while for the Vanilla GAN architecture, single-resolution \textbf{MAT} with hard task weights seems superior, both DCGAN and EDGAN appear to benefit especially from \textbf{MRES-MAT} with uncertainty weights. We also observe cases where adding the auxiliary tasks with a particular weighting strategy massively increases the MMD score (e.g. Toy dataset, GAN-\textbf{MAT UW}). This can happen when the auxiliary task ``overpowers'' the main task, causing the Generator to produce synthetic Moran's I embeddings. This further justifies the generator selection strategy employed throughout our experiments.

\begin{figure}[!ht]
\centering
\includegraphics[scale=0.6]{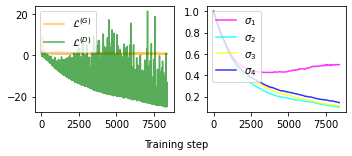}
\caption{Training progression of generator and discriminator losses ($\mathcal{L}^{(G)}, \mathcal{L}^{(D)}$), main task uncertainty $\sigma_{1}$ and auxiliary task uncertainties $\sigma_{2}, \sigma_{3}, \sigma_{4}$ throughout an example training cycle of EDGAN \textbf{MRES MAT UW} using \textit{DEM} training data.}
\label{fig8}
\end{figure}

\textbf{\textit{Task weighting}}: Tables 1 and 2 show that models employing our \textbf{MAT} / \textbf{MRES MAT} auxiliary tasks can reliably produce generators that outperform naively trained models. Uncertainty weight models with \textbf{MRES MAT} represent the ``winning'' Generator in $9$ of $12$ cases. For the hard loss weights, we are always able to find a combination of $\lambda$ and \textbf{MAT} / \textbf{MRES MAT} that outperforms naive training. We thus conclude that while training strategies appear to be highly data and model dependent, one can find a performance-increasing setting in almost all cases. Here, the \textbf{MRES MAT UW} strategy seems to be the safest bet, as it does not require further, manual weight parameter tweaking. Figure 5 shows the losses and learned task uncertainties in an example training cycle of an EDGAN \textbf{MRES MAT UW} model using the \textit{DEM} dataset. More details on all our data, model architectures and training settings can be found in the Appendix. 

\subsection{\textit{Experiment 2:} Predictive spatial modeling}

\begin{figure*}[tb]
\centering
\includegraphics[scale=0.4]{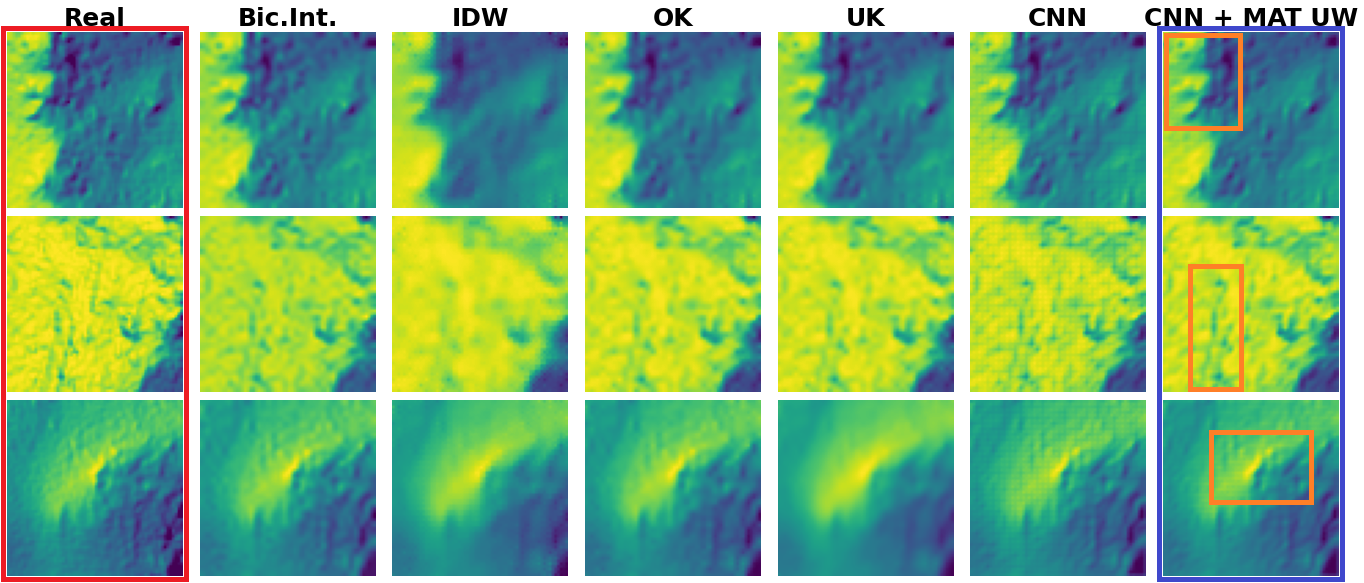}
\caption{Interpolation results on samples from the test set, across the different benchmark models, presenting our CNN + \textbf{MAT UW} model. The orange boxes highlight areas where the improvement over the benchmark models becomes visually apparent.}
\label{fig6}
\end{figure*}

\textbf{\textit{Evaluation}}: In the second experiment, we tackle spatial interpolation, that is, obtaining high-resolution spatial data from a low-resolution input. Spatial interpolation is widely used in real world applications, for example with meteorological measurements \cite{Wang2014}, air quality assessment \cite{Yi2018} or mobile sensing \cite{Nittel2012}. It is a regression task and can be evaluated using the residual mean squared error (RMSE) between real and predicted high-resolution output. As such, this task is comparable to image super-resolution, a popular task in computer vision. Nevertheless, spatial interpolation is particularly focused on reconstructing the spatial patterns of the output. We again train $10$ models per strategy and compare their performance when no model selection is used (final model used for prediction on test set) and when model selection on a validation set is applied, saving the $10$ best models, one from each run.

\textbf{\textit{Data}}: As a common use case in geography and ecology, we use hillshades of DEM data from the National Ecological Observatory Network (NEOS) \footnote{See: \url{https://www.neonscience.org/da-viz-neon-lidar-co13flood-R}} for the interpolation task. The data is pre-processed into a grid of $1674$ $64 \times 64$ tiles (output) and a low-res $32 \times 32$ version (input) by removing every second row and column of the image matrix. For more information on the data used, including summary statistics, please refer to Appendix \textit{A}.

\textbf{\textit{Benchmark Models}}: Spatial statistics provides a range of tools to tackle interpolation problems. Commonly used methods we focus on here are: (1) Bicubic interpolation (\textit{BicInt}), commonly used for interpolating 2-$d$ regular grids, (2) Inverse Distance Weighting (\textit{IDW}) \citep{Shepard1968}, a weighted rolling-average approach, (3) Ordinary Kriging (\textit{OK}) \citep{Cressie1988}, spatial interpolation closely related to Gaussian Process regression and (4) Universal Kriging (\textit{UK}) \citep{Stein1991}, a generalization of OK assuming a polynomial trend model. We compare these established methods to a simple CNN implementation with two hidden layers (5). Again, the modularity of \texttt{SXL} allows us to simply plug-in our \textbf{MAT} by having the CNN interpolate the spatial pattern and its Moran's I embedding from low to high resolution, making the last layers task-specific. In this setting, we do not use the \textbf{MRES MAT}, as our experiments show that further coarsening the already-reduced image is counterproductive. The CNN model main tasks are optimised using MSE loss, while the auxiliary tasks use $\ell 1$ loss. For more details on the experimental setup, please refer to Appendix \textit{B.II}.

\begin{table}
\centering
\resizebox{8cm}{!}{%
\begin{tabular}{l|l|l}
 & \multicolumn{2}{l}{\textbf{RMSE}} \\
\textbf{Model / Task} & \multicolumn{2}{l}{$32 \rightarrow 64$} \\
 & (no model selection) & (model selection)\\
\hline
\hline
BicInt & $0.0667$ & $-$\\
IDW & $0.0693$ & $-$ \\
OK & $0.0801$ & $-$ \\
UK & $0.0796$ & $-$ \\
\hline
CNN & $0.0678 (\pm 0.0128)$ & $0.0503 (\pm 0.0008)$\\
CNN + MAT UW & $\mathbf{0.0649 (\pm 0.0119)}$ & $\mathbf{0.0496 (\pm 0.0006)}$\\
CNN + MAT $\lambda = 0.1$ & $0.0665 (\pm 0.0118)$ & $0.0516 (\pm 0.0018)$ \\
CNN + MAT $\lambda = 0.01$ & $0.0666 (\pm 0.0178)$  & $0.0532 (\pm 0.0033)$    
\end{tabular}}
\caption{Model RMSE scores and their standard deviation on held-out test data for the $32 \rightarrow 64$ interpolation task. The CNN scores are obtained by averaging over $10$ runs each, once taking the final trained model (no model selection) and once selecting the best model according to the validation set (model selection).}
\label{tab:table4}
\end{table}

\textbf{\textit{Findings}}: The results of our experiments are presented in Table 3 and Figure 6. We can again see a positive effect of the \textbf{MAT} on the performance of the CNN model---outperforming all other benchmarks. If no model selection is deployed, both hard loss weights and uncertainty weights produce models that outperform the naively trained CNN. \textbf{MAT UW} models provide the best average performance increase, of around $5\%$. If model selection is utilized, the \textbf{MAT UW} strategy outperforms the naive CNN by about $1.5\%$. Both of these performance increases are statistically significant, according to a paired t-test of the mean prediction scores.

\textbf{\textit{Task weighting}}: Finding the optimal task weighting strategy appears much less tricky as for generative modeling, as the \textbf{MAT UW} strategy prevails in all interpolation experiments, whether model selection is applied or not. We can thus conclude that the \textbf{MAT UW} both consistently and significantly improves training. More details on all our data, model architectures and training settings can be found in the Appendix. 
\section{Conclusion}

In general, our experiments give some insight into the way the auxiliary learning mechanism works, allowing us to compare the method to related ideas in machine learning: First, as mentioned before, we believe the auxiliary tasks to have a regularizing effect on the learning process, preventing models from overfitting on the primary task by forcing them to follow ``spatial rules''. Second, we believe the \textbf{MRES MAT} shares the intuition of moment matching, as we seek to simultaneously minimize the loss of one function at several coarsened resolutions. Third, the \textbf{MRES MAT} also shares the same goal as recent developments in visual self-attention: moving beyond the short distance spatial learning of convolutional layers and accounting for longer distance spatial effects. We also make some empirical observations with respect to the most important design choices of our models: (1) The use of \textbf{MAT} vs. \textbf{MRES MAT} appears to depend on both the input data and the model architecture used, with \textbf{MRES MAT} prevailing in most cases. (2) The optimal auxiliary task weighting strategy varies across generative and predictive modeling experiments, but in most cases uncertainty weights appear to have the edge.

In summary, with \texttt{SXL} we propose the use of single- and multi-resolution measures of local spatial autocorrelation for improving the learning of geospatial processes. We introduce a novel, flexible multi-resolution version of the local Moran's I statistic using coarsened inputs. We demonstrate its integration as an auxiliary task into generative and predictive neural network models, using both hard (static) and task uncertainty (automatically learned) loss weights. We empirically show robust, consistent and significant performance gains of up to ~10\% for generative spatial modeling and up to ~5\% for predictive spatial modeling when using this strategy. We comment on the importance of the exact configuration of the auxiliary tasks, especially choosing single- versus multi-resolution auxiliary tasks and the weighting strategy for auxiliary losses. We see this study further evidence of the importance of integrating domain expertise from GIS into neural network methods for geospatial data, opening a broad range of further research directions: In future work, we plan to expand this idea beyond discrete, regularly gridded data to continuous spatial processes and to assess the applicability of other common spatial metrics for a more comprehensive learning of complex geospatial patterns.

\begin{acks}
This work is supported in part by the \grantsponsor{EPSRC}{UK Engineering and Physical Sciences Research Council} u under Grant No.~ \grantnum{EPSRC}{EP/L016400/1}.
\end{acks}

\bibliographystyle{ACM-Reference-Format}
\bibliography{sigspatial_2021}

\clearpage

\begin{appendix}
    \section*{Appendix}

\subsection*{A: Data Description}

\begin{table*}[ht]
\centering
\begingroup\footnotesize
\begin{tabular}{rrrrrrrrrrr}
 \textbf{Dataset} & $\mathbf{n}$ & \textbf{Min} & $\mathbf{q_1}$ & $\mathbf{\widetilde{x}}$ & $\mathbf{\bar{x}}$ & $\mathbf{q_3}$ & \textbf{Max} & $\mathbf{s}$ & \textbf{IQR} & \textbf{\#NA} \\ 
  \hline
 PetrelGrid & 199680 & 0 & 3 & 7 & 11.2 & 15 & 153 & 12.2 & 12 & 0 \\ 
 DEM (Gen.Mod.) & 1183743 & 76.2 & 246.9 & 327.5 & 331.0 & 406.6 & 886.4 & 117.2 & 159.7 & 1 \\ 
TreeCanopy & 10240000 & 0 & 0 & 31 & 33.8 & 67 & 100 & 34.1 & 67 & 0 \\  
DEM (Sp.Int) & 6856704 & 0 & 126 & 165 & 157.1 & 186 & 254 & 47.6 & 60 & 0 \\ 
  \end{tabular}
\endgroup
\caption{Descriptive statistics of the four real world datasets used for the generative modeling and spatial interpolation experiments.} 
\label{tab:table5}
\end{table*}

Due to a lack of geospatial benchmark datasets within the machine learning community, we run our experiments using one toy dataset and three datasets from real-world geospatial applications.  All data is chosen to represent different spatial patterns and to be closely related to important applications in fields such as climate science or geology.

\paragraph{Toy:} The Toy dataset consists of $32 \times 32$ matrices with values coming from a function creating a Gaussian peak at a random location, which is mirrored diagonally by a Gaussian dip. This function is given as:
\begin{equation}
\begin{aligned}
   f(\mathbf{c_{X}},\mathbf{c_{Y}},s) = 0.75 \exp(-((9\mathbf{c_{X}} - a)^{2} + (9\mathbf{c_{Y}} - b)^{2})/s) \\ -(0.75 \exp(-((9\mathbf{c_{X}} - d)^{2} + 
   (9\mathbf{c_{Y}} - e)^{2})/ s))
\end{aligned}
\end{equation}

where $\mathbf{c_{X}}$ and $\mathbf{c_{Y}}$ are the spatial coordinates mapping the values to the $32 \times 32$ matrix (so in our cases, integers in the range $[0,31]$, $s$ is a positive constant determining the size of the Gaussian peak and dip (we use $s=7$), $a$ and $b$ are random draws from integers in the range $[0,10]$, determining the location of the Gaussian peak and $d = 10 - a$ and $e = 10 - b$ are the location of the Gaussian dip, mirroring the peak diagonally.

\paragraph{PetrelGrid:} The PetrelGrid dataset \citep{Li2013} is composed of geo-referenced seabed relief data. It can be accessed via \textit{R} here: \url{https://rdrr.io/cran/spm/man/petrel.grid.html}

\paragraph{DEM (Generative Modeling):} We use two different digital elevation model (DEM) based datasets, one for the generative modeling experiments and one for the spatial interpolation experiments. The DEM for generative modeling is chosen as it is rather small, enabling us to assess how our proposed method deals with data scarcity. An applicably small DEM dataset, providing a DEM of the area surrounding Lake Sunapee, NH, USA can be found as part of the \texttt{elevatr} \textit{R} package; accessible via: \url{https://rdrr.io/cran/elevatr/man/lake.html}.

\paragraph{TreeCanopy:} This dataset contains data on global forest coverage. We use tree canopy, which describes canopy closure for all vegetation taller than 5m in height. The data comes from the University of Maryland's "Global Forest Change" project \citep{Hansen2013}, documenting the global loss of forests in the light of climate change and forest exploitation. Specifically, we use data within the geographic area 50-60N / 100-110W; an area lying in continental Canada and representing a broad range of forest coverage types. The data can be accessed via: \url{http://earthenginepartners.appspot.com/science-2013-global-forest/download_v1.6.html}.

\paragraph{DEM (Spatial Interpolation):} The second, larger DEM dataset used for the spatial interpolation experiments is part of a LiDAR data collection conducted by the National Ecological Observatory Network (NEOS). Specifically, we use DEM hillshades from the NEOS training exercise outlined here: \url{https://www.neonscience.org/da-viz-neon-lidar-co13flood-R}. Hillshades are used to visualize terrain as shaded reliefs, where shades depend on a (synthetic) light source (e.g. the sun shining at a modelled angle).  

All our data is processed into regular grids of either size $32 \times 32$ or $64 \times 64$. For the exact processing of each of the datasets, please refer to our code.

\subsection*{B: Experimental Setting, Model Architectures and Compute}

\noindent\textbf{B.I: Generative Spatial Modeling}

\textit{Setup:} Our main experimental findings, the MMD scores displayed in Table 1 and Table 2 (main paper), are obtained from training generative models on $60\%$ of the data, holding out $20\%$ of the data for validation and model selection, and $20\%$ for computing the displayed test scores. This setting is used for all four experimental datasets. For each dataset, model architecture and auxiliary task setting, we train $10$ GANs with different random initializations, in each cycle saving the best generator according to tests on validation data. We then choose the best out of the $10$ trained generators (again according to the validation score) to compute test scores.

\textit{Model Architecture and Optimization:} Here we briefly describe the model architectures of the different generative models used in the experiments working with $32 \times 32$ inputs (the models for the $64 \times 64$ input are adapted to fit the larger input). For the implementation of these models, please refer to our code.

The Vanilla \textbf{GAN} architecture used consists of a Generator with four hidden linear layers, supported by Leaky ReLU and $1d$ BatchNorm layers. The Discriminator has two hidden linear layers supported by Leaky ReLU layers and one linear task-specific layer.

The \textbf{DCGAN} architecture used consists of a Generator with a linear initialization layer, followed by three hidden (de-)convolutional layers, supported by ReLU and $2d$ BatchNorm layers. The Discriminator contains two convolutinal layers supported by Leaky ReLU and $2d$ BatchNorm layers, followed by one task-specific convolutional layer with a final linear transformation. For more information on DCGAN, please refer to the original publication \citep{Radford2016}.

The \textbf{EDGAN} architecture used consists of an Encoder-Decoder Generator, where the Encoder contains three convolutional layers, supported by Leaky ReLU and $2d$ BatchNorm layers and the Decoder contains three (de-)convolutional layers supported by ReLU layers. The Discriminator has five hidden convolutional layers suppoerted by Leaky ReLU and $2d$ BatchNorm layers, followed by a last, task-specific convolutional layer. For more information on the EDGAN architecture, please refer to \citep{Zhu2019}, the study which motivated the use of this benchmark.

\textit{Model Training:} All models are trained using the binary cross entropy criterion to compute losses. Optimization through backpropagation is conducted using the Adam algorithm with a learning rate of $0.001$ and $\beta$ values of $[0.5, 0.999]$. Experiments with the \textit{Toy} dataset run for $40$ epochs, with the \textit{PetrelGrid} dataset for $500$ epochs, with the \textit{DEM} dataset for $100$ epochs and with the \textit{TreeCanopy} dataset for $100$ epochs. All training is conducted on GPUs provided via \textit{Google Colab}, which includes \textit{Tesla K80, Tesla T4 and Tesla P100} GPUs. The model training times do not exceed 30 minutes at the longest.

\textit{Evaluation}: To evaluate our models, we generate synthetic data from the different types of GANs and compare how faithful the generated samples are compared to true samples. To assess model quality, we use the Maximum Mean Discrepancy (MMD) metric \citep{Borgwardt2006}, a distance measure between distributions based on mean embeddings of the features. For distributions $P$ and $Q$, the MMD is defined as $MMD(P,Q) = || \mu_{P} - \mu_{Q} ||_{\mathbb{R}^{d}}$. The empirical MMD for random variables $x_{i}$ and $y_{i}$ of length $n$ is given as

\begin{equation}
\begin{aligned}
    \widehat{MMD}^{2} = \frac{1} {n(n - 1)} \sum_{i \neq j} k(x_{i},x_{j}) +  \\
    \frac{1}{n(n - 1)} \sum_{i \neq j} k(y_{i},y_{j}) -  \frac{2}{n^{2}} \sum_{i,j} k(x_{i},y_{j}),
\end{aligned}
\end{equation}

where $k: \mathcal{X} \times \mathcal{X}$ represents a positive-definite kernel---in our case a radial basis function (RBF) kernel. The more similar the data distributions $P$ and $Q$ are, the closer the MMD metric gets to $0$. \\

\noindent\textbf{B.II: Predictive Spatial Modeling}

\textit{Setup:} Our main experimental findings for the spatial interpolation experiments, the RMSE scores displayed in Table 3, are obtained from training the CNN models on $60\%$ of the data, selecting the best model using $20\%$ and finally computing the scores on held-out $20\%$ held-out test data. This is done ten times and the test scores are then averaged. Note that the non-neural network based benchmark models (bicubic interpolation, IDW, and kriging) do not require training; rather inference is made directly on the testing samples.

\textit{Model Architecture and Optimization:} We use a simple CNN for the predictive modeling experiments. It consists of three convolutional layers, supported by ReLU and $2d$ BatchNorm layers. When applying the auxiliary tasks to the model, the last two convolutional layers are made task-specific. Please refer to our code for the exact implementation of the models.

\textit{Model Training:} All models are trained using the mean squared error (MSE) criterion to compute losses. Optimization through backpropagation is conducted using the Adam algorithm with a learning rate of $0.001$ and $\beta$ values of $[0.5, 0.999]$, running for $150$ epochs. All training is conducted on GPUs provided via \textit{Google Colab}, which includes \textit{Tesla K80, Tesla T4 and Tesla P100} GPUs. The individual model training times do not exceed 15 minutes at the longest.

\textit{Evaluation}: The final evaluation scores on held-out test data are computed as the root mean squared error (RMSE) between real values $y_{i}$ and predicted values $\hat{y}_{i}$ of length $n$:
\begin{equation}
    RMSE(y_{i},\hat{y}_{i}) = \sqrt{\frac{1}{n} \sum (y_{i} - \hat{y}_{i})^{2}}
\end{equation}

\subsection*{C: Code}

All our experiments are implemented using \textit{PyTorch}\footnote{\url{https://pytorch.org/}}. The code for our experiments can be found here: \url{https://drive.google.com/file/d/1ShOBV7RifMdS9LYsySOM084KL5aD-wkm/view?usp=sharing}.
\end{appendix}

\end{document}